\documentclass[letterpaper, 10 pt, conference]{ieeeconf}  

\IEEEoverridecommandlockouts                              

\usepackage{amsmath,amssymb,amsfonts}
\usepackage{xcolor}
\usepackage{url}
\usepackage{siunitx}
\usepackage{graphicx}
\usepackage{dblfloatfix}
\usepackage{float}
\graphicspath{ {./figures/} }
\usepackage{cite}
\usepackage{subfigure}

\usepackage{textcomp}

\usepackage{hyperref}

\usepackage{bm}
\usepackage{float}
\usepackage[font=small,skip=7pt]{caption}
\usepackage[linesnumbered,noend,ruled,vlined]{algorithm2e}
\usepackage{microtype}
\usepackage{multirow}
\usepackage{graphicx}
\usepackage{rotating}
\SetKwInput{KwInput}{Input}                
\SetKwInput{KwOutput}{Output}              
\newlength\fwidth
\usepackage{mathtools}

\usepackage{siunitx}
\usepackage{epstopdf}
\usepackage{mathtools}
\usepackage[T1]{fontenc}
\DeclareGraphicsExtensions{.pdf,.eps}
\usepackage{textcomp}
\usepackage[ISO]{diffcoeff}
\usepackage{caption}
\usepackage{amsmath}

\usepackage{titlesec}
\titlespacing*{\section} {0pt}{6pt}{6pt}
\titlespacing*{\subsection} {0pt}{4pt}{4pt}

\def\BibTeX{{\rm B\kern-.05em{\sc i\kern-.025em b}\kern-.08em
    T\kern-.1667em\lower.7ex\hbox{E}\kern-.125emX}}

\usepackage[acronym,nomain,nonumberlist]{glossaries}
\usepackage{url}

\title{\LARGE \bf

A Hierarchical Graph-Based Terrain-Aware Autonomous Navigation Approach for Complementary Multimodal Ground-Aerial Exploration}

\author{Akash Patel, Mario A.V. Saucedo, Nikolaos Stathoulopoulos, Viswa Narayanan Sankaranarayanan, \\ 
Ilias Tevetzidis, Christoforos Kanellakis, and George Nikolakopoulos
\thanks{The authors are with Robotics \& AI Team, Department of Computer, Electrical and Space Engineering, Lule\r{a} University of Technology, Lule\r{a} SE-97187, Sweden. Corresponding author: \tt\small akapat@ltu.se}
\thanks{This work has been partially funded by the European Unions Horizon 2020 Research and Innovation Programme under the Grant Agreement No. 101138451 PERSEPHONE.}%
}

\begin{document}

\maketitle
\thispagestyle{empty}
\pagestyle{empty}

\begin{abstract}

Autonomous navigation in unknown environments is a fundamental challenge in robotics, particularly in coordinating ground and aerial robots to maximize exploration efficiency. This paper presents a novel approach that utilizes a hierarchical graph to represent the environment, encoding both geometric and semantic traversability. The framework enables the robots to compute a shared confidence metric, which helps the ground robot assess terrain and determine when deploying the aerial robot will extend exploration. The robot's confidence in traversing a path is based on factors such as predicted volumetric gain, path traversability, and collision risk. A hierarchy of graphs is used to maintain an efficient representation of traversability and frontier information through multi-resolution maps. Evaluated in a real subterranean exploration scenario, the approach allows the ground robot to autonomously identify zones that are no longer traversable but suitable for aerial deployment. By leveraging this hierarchical structure, the ground robot can selectively share graph information on confidence-assessed frontier targets from parts of the scene, enabling the aerial robot to navigate beyond obstacles and continue exploration.

\end{abstract}


\section{Introduction}\label{sec:introduction}

In the field of robotic exploration, ground and aerial robots have proven their pioneering capabilities to explore challenging terrains in underground deployment. Ground robots, such as wheeled robots and legged robots, are known for their extended operational endurance. On the other hand, aerial robots, such as drones are particularly interesting in 3D voids and vertical passages-like structures. The aerial robots have lower endurance limit compared to the ground robots; however, they can operate in situations where ground robot are simply unable to continue, for example, flying over obstacles. Inspired from the lessons learned during the DARPA subterranean challenge~\cite{chung2023into} \cite{ebadi2023present}, this article aims at laying a foundation for a combined mobility robotic system in which the ground robot can identify deployment points using multimodal traversability information, encoded into navigation graphs. There exist multiple exploration architectures tailored for ground and aerial robots \cite{dang2020graph} \cite{dharmadhikari2020motion} \cite{yang2021graph} deployed in the DARPA subterranean challenge. Our previous work on STAGE planner \cite{patel2024stage} focuses on graph based exploration approach tailored to solve dynamic uncertainty in the explored map while re-routing robot only using graph updates. 

\begin{figure}[h]
    \centering
        \includegraphics[width=\linewidth]{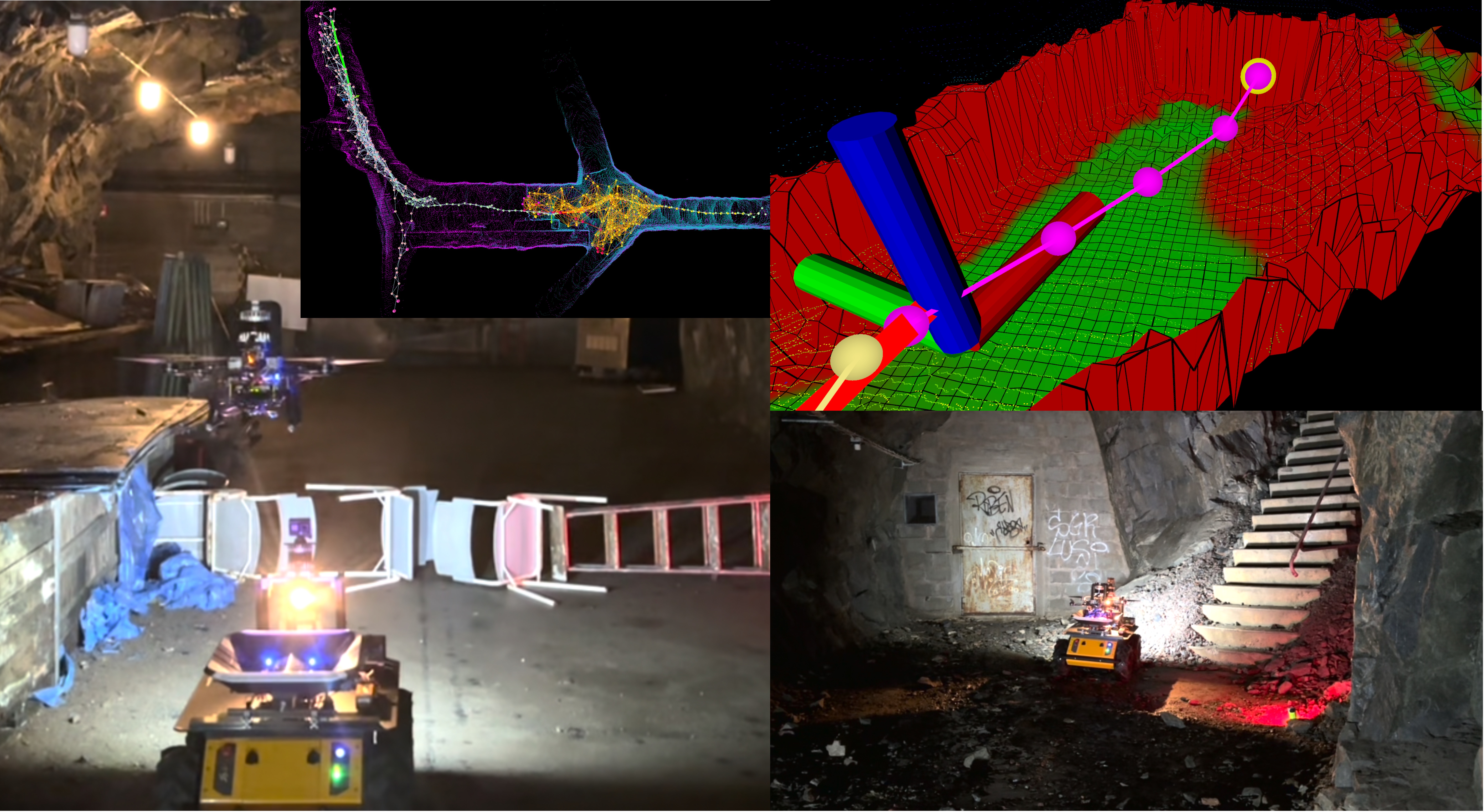}
    \caption{The complementary ground-aerial robotic system collaboratively explores a challenging subterranean environment. The ground robot autonomously identifies deployment points, based on multi-modal traversability and encoded within hierarchical graphs, enabling continued aerial exploration.}
    \label{fig:marsupial}
\end{figure}

In this article we leverage a graph based approach, as a baseline exploration framework, while incorporating a novel strategy that maintains hierarchy of graphs, encoding geometric and semantic traversability for the ground robot to assess exploration paths. The literature shows number of works attempting at ground-aerial robotic deployment in exploration scenarios. Lindqvist et al. in \cite{lindqvist2022multimodality} presented a robotic system combining legged and aerial mobility for search and rescue operation. The legged robot carries the aerial robot to a manually defined deployment point, followed by releasing the aerial robot to explore using a reactive tunnel following method. This method particularly lacks in autonomy for a collaborative exploration mission, since the deployment point is manually defined by an operator and no knowledge is shared between the robots for cooperative exploration. De Petris et al. in \cite{de2022marsupial} presented the work of team CERBERUS \cite{tranzatto2022cerberus} in DARPA subterranean challenge work on marsupial walking-flying robotic exploration. In this method, when the legged robot reaches local exploration completion, it identifies partially seen areas that can be beneficial from aerial exploration. This approach also utilizes a graph based exploration planner \cite{dang2020graph} for the legged robot to maintain a graph composed of sparsely sampled vertices connected with robot's poses and frontier vertices. The legged robot identifies frontier in blocked part of the tunnel and shares explored map with aerial robot to start exploration from the shared frontier. While this approach \cite{de2022marsupial} presents a higher level of autonomy for the collaborative mission, it relies on sharing explored map and manually co-localizing the robots. Furthermore, the ground robot only relies on occupancy map information to depict the barrier as an untraversable area meaning the negative obstacles (steep stairs, or cavities in the ground) would not be identified as potential deployment zone for aerial robot. On the application side of the ground-aerial collaborative exploration, the work presented in \cite{butzke20153} and \cite{shen2017collaborative} presents the missions for target identification and localization while utilizing navigation capabilities of heterogeneous robotic system.

The related works on ground-aerial cooperative exploration studies shows that the heterogeneous robotic systems mostly rely on pre-defined scenarios to identify when aerial robot should be deployed and what level of information should be shared between the robots. In contrast to that, the proposed method incorporates multimodal traversability and frontier information to construct a robust confidence metric that helps the ground robot to autonomously identify deployment zone. On the other aspect of multi robot collaboration, the proposed approach allows the ground robot to have control over what level of minimal information needs to be shared with aerial robot for rapid deployment without requiring to share the full map. An example of fully autonomous ground-aerial collaborative exploration using the proposed approach is shown in Fig. \ref{fig:marsupial}.


\section{Contributions}

Building on the limitations of current state-of-the-art ground-aerial collaborative exploration methods, the proposed approach tackles key challenges and advances the autonomy of complementary robotic systems through the following novel contributions.

The first contribution of this work introduces a novel approach that incorporates geometric and semantic traversability information to construct a hierarchical graph representing the robot's understanding of the terrain. The approach integrates 3D LiDAR data into a high resolution grid structure to define geometric characteristics, such as elevation, slope, and surface roughness, within a 2.5D map. Additionally, it offers a fundamental improvement in voxelized representation within 3D occupancy maps by integrating semantic traversability values directly into the voxel state. Vertices along the exploration path are projected onto the 2.5D map to query geometric traversability, while ray casting is used on the 3D voxel map to query semantic traversability, allowing the robot to calculate its confidence in traversing a particular path. Based on this confidence metric the ground robot identifies untraversable zones where the ground-aerial robotic system could benefit from the aerial deployment to access the inaccessible zones. 

The second contribution introduces an ad hoc information-sharing mechanism between the ground and aerial robots for high-level knowledge transfer without the need to share explored maps. By utilizing a hierarchical graph for navigation, the ground robot can directly incorporate the frontiers that can benefit from aerial exploration into a shared graph. This allows for an ad-hoc dynamic graph sharing with the aerial robot without the need of sharing explored map. As part of the integration of autonomy modules for real life subterranean exploration, we also open source the code for robotics community to leverage ROS2 implementation of traversability integrated occupancy mapping. 

By integrating the aforementioned multimodal traversability, hierarchical graph, and ad hoc ground-aerial coordination strategies, this paper presents a fully autonomous exploration approach for subterranean environments, demonstrating the combined use of a heterogeneous robotic system.

\section{Methodology } \label{sec:methodology}

 The proposed ground-aerial robotic exploration problem is broken down into three sub problem. A) Ground robot exploration and untraversable zones identification. B) Ground-aerial robot coordination and knowledge sharing and C) Aerial exploration after deployment from the ground robot. In order to achieve a fully autonomous ground-aerial collaborative exploration, we present the hierarchical graph based multi-modal traversability aware exploration approach in the following subsections. 

\subsection{Geometric Traversability} \label{subsec:geometric_traversability}

\begin{figure*}[htbp]
    \centering
        \includegraphics[width=\linewidth]{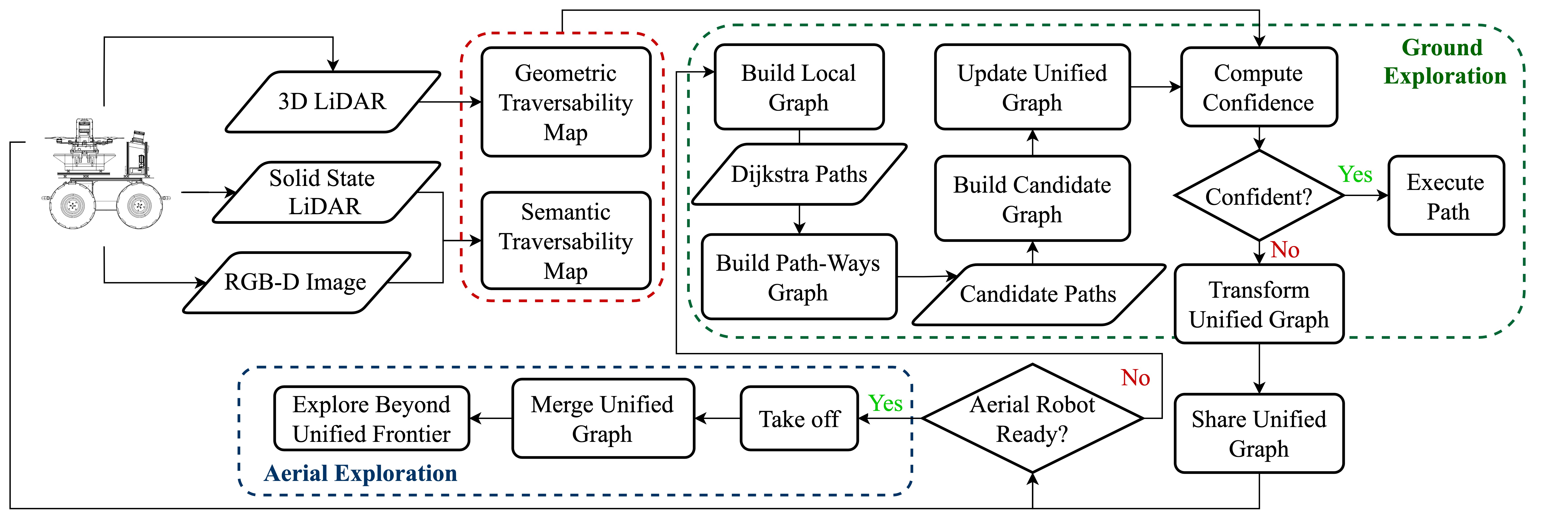}
    \caption{The proposed ground-aerial collaborative exploration algorithm flow chart}
    \label{fig:flow_chart}
    \vspace{-1.2em}
\end{figure*}

Let us denote the state vector of ground robot as $\zeta_{G}$ where $\zeta = \{(x_{i}, y_{i}, z_{i}, \psi_{i})\ |\ i \in N \}$ represents representing the robot's 3D position and Heading ($\psi$). The instantaneous point cloud scans of the ground robot is denoted as $\mathcal{P}_{G}$  where $\mathcal{P} = \{ (x_{i}, y_{i}, z_{i})\ |\ i \in N \}$. Utilizing the approach described in \cite{fankhauser2014robot} we construct a robot-centric local elevation map $\mathbb{M}^{E}$ centered around ground robot's state $\zeta_{G}$ and aligned with the ground robots heading. Derived from this high resolution (1-5 cm) elevation map $\mathbb{M}^E$, and similar to the approach presented in \cite{chilian2009stereo}, we calculate the terrain risk factor $\ \mathclap{R}\ $ based on slope $s$, surface roughness $r$ and step height $h$ as below. 
\begin{align}
    \mathclap{R}\  =  w_{s} \frac{s}{s_{crit}} + w_{r} \frac{r}{r_{crit}} + w_{h} \frac{h}{h_{crit}}
\end{align}

Where, $w_s$, $w_r$ and $w_h$ are weights associated with slope, surface roughness and step height that sum up to 1. In the proposed approach we formulate the geometric traversability as an inverse linear function of the risk factor defined as ${T}^{g} = 1 - \mathclap{R}\ $ where the geometric traversability ${T}^{g} \in  [0, 1]$. The critical values of slope ($s_{crit}$), roughness ($r_{crit}$) and step height ($h_{crit}$) are defined as hard constraints based on robot specific operational values, exceeding the critical values of slope, roughness or step height would result in setting 0 traversability for a cell. The ground robot size is defined based on a bounding box encapsulating the robot dimensions. The bounding box is denoted as ${B}_{G} = \{ \zeta, l, w, d \}$ where $l$, $w$ and $d$ represent the length, width and height of the 3D cuboid centered around robot state $\zeta_{G}$ encapsulating the ground robot. In order to reduce computational overhead we only want to query the cells of high resolution map that capture the projected robot size onto the 2.5D traversability map $\mathbb{M}^{T}$. We achieve this by fitting a polygon $P$ based on robot bounding box and projecting it onto the traversability map $\mathbb{M}^{T}$. Given the known half length $l_h = \frac{l}{2}$ and half width $w_h = \frac{w}{2}$ of the bounding box we can define a generic equation to get the vertex position $\mathbf{p}_{i}$ of the polygon as:

\[
\mathbf{p}_{(x,y)} = (\zeta^{x}, \zeta^{y}) + \mathbf{R}(\zeta^{\psi}) \cdot \begin{pmatrix} s_1 l_h \\ s_2 w_h \end{pmatrix}
\]
where $s_1, s_2 \in \{-1, 1\}$, and $\mathbf{R}(\zeta^{\psi})$ is the rotation matrix:


Let ${T}^{g}(i,j) \in [0, 1]$ represent the geometric traversability value of the grid cell at position $(i,j)$ in the geometric traversability map $\mathbb{M}^{G}$. The set of cells inside the polygon $P$ is denoted by:
\[
\mathcal{C}(P) = \left\{ (i,j) \mid (x_{i,j}, y_{i,j}) \in P \right\}
\]
where $(x_{i,j}, y_{i,j})$ are the coordinates of the center of the cell $(i,j)$.

The set of geometric traversability values inside the polygon is:
\[
{T}^{g}_{P} = \left\{{T}^{g}(i,j) \mid (i,j) \in \mathcal{C}(P) \right\}
\]

The average geometric traversability ${T}^{g}_{\text{avg}}$ of the grid cells inside the polygon $P$ is defined as:
\[
{T}^{g}_{\text{avg}} = \frac{1}{|\mathcal{C}(P)|} \sum_{(i,j) \in \mathcal{C}(P)}{T}^{g}(i,j)
\]
where $|\mathcal{C}(P)|$ is the number of cells inside the polygon $P$, and ${T}^{g}(i,j)$ is the geometric traversability value of the cell $(i,j)$.

\subsection{Semantic Traversability} \label{subsec:semantic_traversability}

In this article, we use a modified version of the architecture proposed in \cite{saucedoEAT} for traversability-based terrain classification, where a deep learning architecture is leveraged for terrain smoothness estimation based on RGB images.
The overall segmentation process considers 4 traversability levels:
    \textit{Untraversable}. All terrain types that are considered untraversable for the platform under consideration. This refers to objects like walls and other obstacles (e.g. rocks, leftover material, etc.).
    \textit{Undesirable}. The least traversable terrain, mainly due to its dangerous nature, like is the case of water or ice.
    \textit{Rough}. Terrain that is traversable but not optimal in contrast to better alternatives. Some examples are gravel and sand.
    \textit{Optimal}. Terrain that is fully traversable, asphalt and dirt will be two of the most common examples.

The resulting semantic segmentation is then fused with the surface normals estimated from depth images using a traversability decay function:
\begin{equation}
     \boldsymbol{T}^s(\alpha,\ \theta) = \alpha \exp ^ {-\frac{\theta}{\alpha}}
    \label{eqn:trav}
\end{equation}

where $T^s \in [0,1]$ is the traversability score, with the higher values (i.e. the closer to 1) representing the more traversable terrain and vice versa for the lower values, while $\alpha$ is the terrain label and $\theta$ is the slope estimated from the surface normals. The function behaves so that based on the terrain class the traversability decays faster in function of the slope, for example, dirt becomes non-traversable at less steeped slopes than asphalt. In addition, the maximum traversability value that the function can take depends on the terrain class. Since even if both, dirt and asphalt, are perfectly flat, the asphalt is still a most desirable terrain to traverse.

Finally, the estimated traversability scores are map to the robots point cloud $\mathcal{P}$ using the camera intrinsic parameters. Let the world frame $\mathbb{W}$ be fixed and define as the workspace of the robotic platform, the LiDAR frame $\mathbb{L}$ be located on the LiDAR sensor, while the camera frame $\mathbb{C}$ is attached on the camera sensor. The projection of a point $p^\mathbb{L} = (x, y, z)$ to Image coordinates is performed using the homogeneous transformation matrix:
\begin{equation}
    \begin{bmatrix}
     u\\
     v\\
    1
    \end{bmatrix}
    =
    \underbrace{
     \begin{bmatrix}
     f_x&  0&  c_x\\
     0&  f_y&  c_y\\
     0&  0&  1\\
    \end{bmatrix}
    }_K
    \underbrace{
    \begin{bmatrix}
     r_{11}&  r_{12}& r_{13}& t_{x}\\
     r_{21}&  r_{22}& r_{23}& t_{y}\\
     r_{31}&  r_{32}& r_{33}& t_{z}\\
    \end{bmatrix}
    }_{[\mathbf{R}|t]}
    \begin{bmatrix}
     x\\
     y\\
     z\\
    1
    \end{bmatrix}
    \label{eq:proj}
\end{equation}

\begin{figure*}[t]
    \centering
        \includegraphics[width=\linewidth]{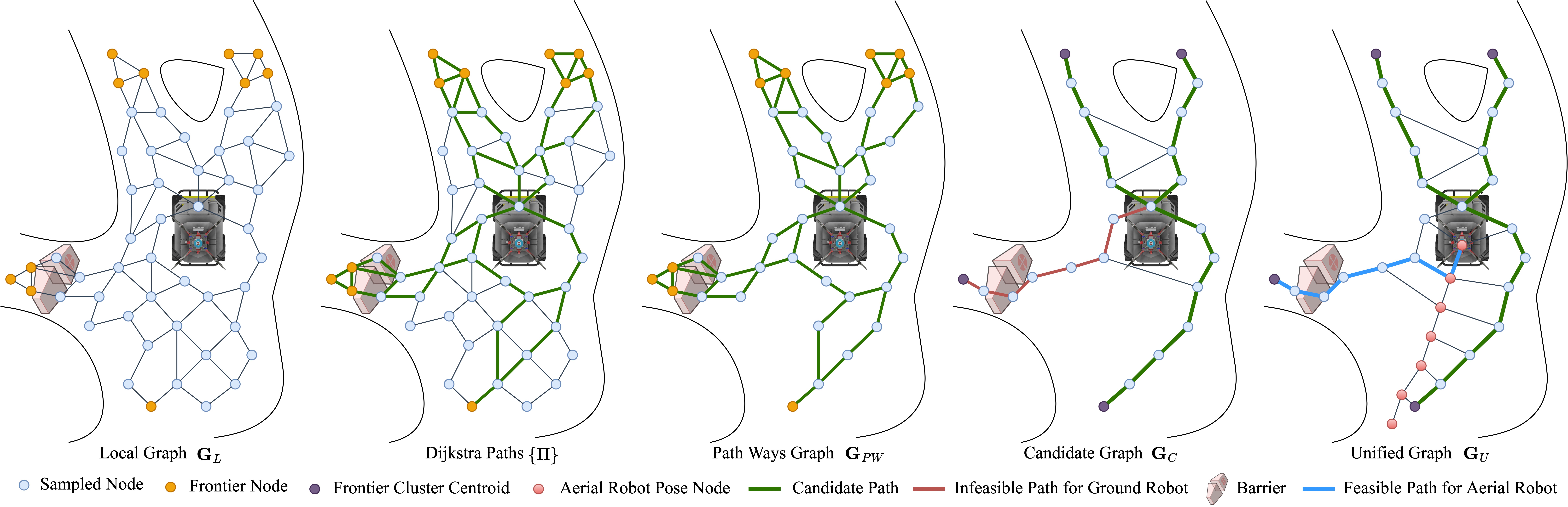}
    \caption{The proposed hierarchical decomposition of navigation graphs}
    \label{fig:concept}
    \vspace{-1.2em}
\end{figure*}

where the vector $[x, y, z,1]^T$ represents the 3D coordinates of any given point on the point cloud, $\mathbf{R}$ and $t$ are the rotation matrix and the translation vector, respectively, between the camera frame $\mathbb{C}$ and the LiDAR frame $\mathbb{L}$, $f_x$ and $f_y$ are the focal lengths of the camera with principal point $(c_x,c_y)$, and $p^\mathbb{I} = (u,v)$ is the projection of the point $p^\mathbb{L}$ to 2D image coordinates.
Likewise, the transformation of a point $p^\mathbb{L}$ to $\mathbb{W}$ is performed using the homogeneous transformation matrix $\mathbf{T}^{\mathbb{W}}_{\mathbb{L}}$. In this work, we utilize Direct Lidar Inertial Odometry \cite{chen2023direct} for state estimation and to compute the transformation matrix $\mathbf{T}^{\mathbb{W}}_{\mathbb{L}}$ from robot odometry and a known LiDAR to robot body frame transform.  

This gives as a result the labeled traversability point cloud $\mathcal{P}^{T^s}$ encompassing the semantic traversability for all points in $\mathcal{P}$:
\begin{equation}
    \mathcal{P}^{T^s} = \{ (x_{i}, y_{i}, z_{i}, T_{i}^s)\ |\ i \in N \}
\end{equation}

In a classical volumetric map representation, each voxel $v$ can have one of the three distinct states denoted as $v^{occ}$, $v^{fre}$ and $v^{unk}$, corresponding to occupied, free and unknown states. To build the 3D volumetric map $\mathbb{M}^{V}$ of the environment, we use modified version of the voxblox mapping architecture presented in \cite{oleynikova2017voxblox}. In the proposed approach, we leverage the previously computed semantic traversability point cloud $\mathcal{P}^{T^s}$ as an input to the Truncated Signed Distance Fields (TSDF) integrator. Each TSDF voxel in the $\mathbb{M}^{V}$ incorporates an additional attribute called semantic traversability. The traversability field from $\mathcal{P}^{T^s}$ is only mapped into the occupied voxel $v^{occ}$ since the primary focus for the ground robot is to analyze the terrain traversability. Similar to the approach in \cite{oleynikova2017voxblox}, the layers of each TSDF voxel contain discrete blocks that are indexed by their position in the global volumetric map $\mathbb{M}^{V}$. For recursive traversability query in a high resolution voxel map, this integration approach is particularly interesting because the mapping between TSDF block position and it's memory location is stored in a hash table. This allows for $\mathcal{O}(1)$ update and look-up complexity compared to the classical octree query which is generally $\mathcal{O}(\log n)$. 

Let $T^{s}(v_{i,j,k}) \in [0, 1]$ denote the semantic traversability value of an occupied voxel $v_{i,j,k}^{\text{occ}}$ at index $(i,j,k)$.

The set of occupied voxels whose centers lie inside the polygon $P$ is given by:
\[
\mathcal{V}_P^{\text{occ}} = \left\{ v_{i,j,k}^{\text{occ}} \mid (x_{i,j}, y_{i,j}) \in P \right\}
\]
where $(x_{i,j}, y_{i,j})$ are the 2D coordinates of the voxel centers projected onto the plane.

The average semantic traversability $T^{s}_{\text{avg}}$ of the occupied voxels inside the polygon is:
\[
T^{s}_{\text{avg}} = \frac{1}{|\mathcal{V}_P^{\text{s}}|} \sum_{v_{i,j,k}^{\text{occ}} \in \mathcal{V}_P^{\text{occ}}} T^{s}(v_{i,j,k})
\]

\subsection{Hierarchical Graph Representation} \label{subsec:hierarchical_graphs}


Insights from the DARPA Subterranean Challenge have demonstrated that graph-based path planning \cite{dang2020graph} is an effective approach for navigating such environments. Building on our previous work with the STAGE planner \cite{patel2024stage}, we propose a novel hierarchical navigation graph structure designed for exploration and path planning within a ground-aerial robotic system.

\begin{figure*}[t]
    \centering
        \includegraphics[width=\linewidth]{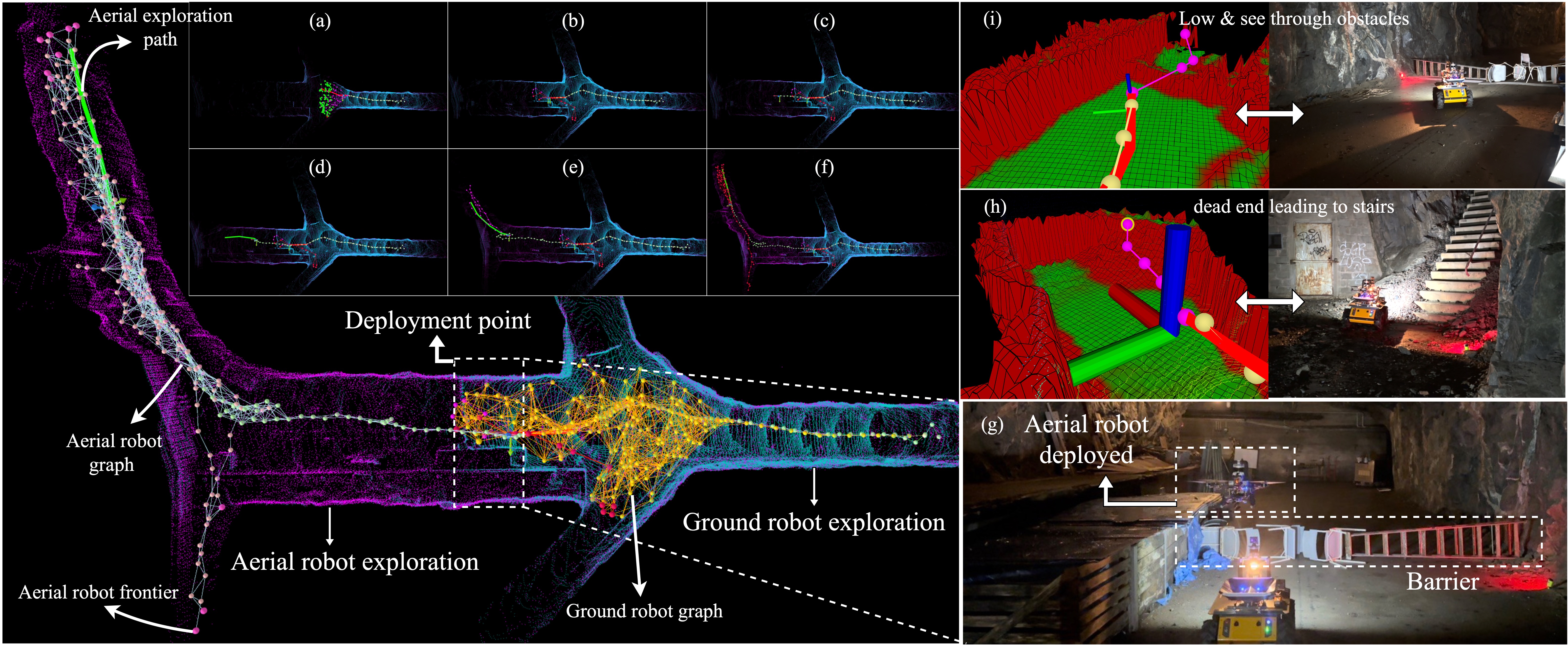}
    \caption{The fully autonomous collaborative ground-aerial robotic exploration of a subterranean environment. Snapshots from field experiments highlighting different stages of exploration and key moments leading to aerial deployment.}
    \label{fig:experiment_collage}
    \vspace{-1.2em}
\end{figure*}

Using the robot state $\zeta_{G}$ as root, the set of nodes $\mathcal{N} = \{ n_{1}, n_{2}, .. , n_{n} \}$ where, $n_{i} = \{x_{i}, y_{i}, z_{i}, \psi_{i}\}$ are sampled within the local map $\mathbb{M}_{L}$, $(\mathbb{M}_{L} \subset \mathbb{M}^{V})$. The sampled nodes are interconnected using \textit{kdtree} based nearest neighbour search to form the set of collision free edges $\mathcal{E} = \{ e_{1}, e_{2}, .. , e_{n} \}$ that represent a dense local graph $\mathbf{G}_{L} = (\mathcal{N}, \mathcal{E})$. We use the local graph as parent graph in the hierarchy structure and all other subsequent graphs are derived from the $\mathbf{G}_{L}$. Each node in the local graph is evaluated to compute predicted volumetric gain based on \eqref{eqn:volumetric_gain} where, $\hat{u}$, $\hat{f}$ and $\hat{o}$ represent normalized values of occupied, free and unknown voxels within the robot's field of view frustum. $w_{u}$, $w_{f}$ and $w_{o}$ represent the scalar weights for unknown, free and occupied voxels. 
\begin{equation}\label{eqn:volumetric_gain}
    \Phi (n_{i}) = 
    {\log\left( \frac{w_{u} e^{\hat{u}} + w_{f} e^{\hat{f}}}{w_{o}  e^{\hat{o}}} \right)}
\end{equation}

Nodes with a volumetric gain higher than a threshold value $\Phi_{min}$ are marked as frontier nodes in the local graph $\mathbf{G}_{L}$. Using Dijkstra's shortest path algorithm on $\mathbf{G}_{L}$, we compute the shortest paths from the robot's state $\zeta_{G}$ to each frontier node. Each path consists of a sequence of connected nodes $n_i \in \mathcal{N}$, starting from the robot's position and ending at a frontier node. By reusing the nodes from the Dijkstra paths, we construct a pathways graph $\mathbf{G}_{PW}$ by refining the connections (edges) between nodes. Since $\mathbf{G}_{PW}$ is derived from the nodes along the Dijkstra paths, it represents a sparser version of $\mathbf{G}_{L}$. Dijkstra's shortest path algorithm is reapplied on the path-ways graph, and the resulting paths are compared using Dynamic Time Warping (DTW) distances to identify candidate paths connecting the robot's position to the clustered frontier nodes. The path-ways graph is further reduced to form the candidate graph $\mathbf{G}_{C}$ by reusing only the nodes along these candidate paths. A visual representation of the proposed hierarchical graph structure is presented in Fig. \ref{fig:concept}.
Each path $\Pi = \{n_{1}, n_{2}, \dots , n_{i}\}$ is evaluated to compute confidence score based on the multi-modal (geometric \& semantic) traversability for the ground robot. The confidence score $C(n_{i})$ for each node along the path $\Pi$ is computed using a sigmoid function as shows in \eqref{eqn:node_confidence_score}. 
\begin{align}\label{eqn:node_confidence_score}
    C(n_i) = \frac{1}{1 + e^{-\left( w_g \cdot T^{g}_{\text{avg}} + w_s \cdot T^{s}_{\text{avg}} + w_v \cdot \Phi(n_i) \right)}}
\end{align}

Where, $w_g$, $w_s$ and $w_v$ represent scalar weights for geometric, semantic traversability and volumetric gain respectively. Each node on the candidate path is evaluated by projecting the polygon $P$, whose centroid corresponds to the coordinates of node $n_{i} \in \Pi$, onto the 2.5D map to assess average traversability $T_{avg}$.
\begin{equation}
\Pi_{C} =
\begin{cases} 
    \frac{\sum_{i=0}^{k} C(n_{i})}{k}, & \text{if } C(n_{i}) > C_{\text{crit}} \, \forall i \\
    e^{-\lambda} \cdot \frac{\sum_{i=0}^{k} C(n_{i})}{k}, & \text{if } \exists i \, \text{such that } C(n_{i}) \leq C_{\text{crit}}
\end{cases}
\label{eqn:path_confidence_score}
\end{equation}

The confidence score of a candidate path is computed based on \eqref{eqn:path_confidence_score}. To enhance the robustness of deployment zone identification, we impose a strict confidence threshold, $C_{\text{crit}}$. If any node along the candidate path has a confidence score lower than this critical value, the overall confidence of the path is penalized by a factor of $e^{-\lambda}$, where $\lambda$ is a tunable scalar penalty parameter, ensuring that paths with uncertain or risky sections are de-prioritized. This formulation offers the advantage of robustly identifying potential aerial deployment zones, as discussed in Section \ref{subsec:deployment_identification}.


\section{Experimental Evaluation}

The proposed framework was field-tested using a ground robot (Clearpath Husky A200) and a custom-built aerial robot (Shafter), combined to form a ground-aerial robotic platform. We evaluated the framework based on two key metrics. The first metric assesses the ground robot’s ability to identify untraversable zones (areas with low confidence for traversing) that could serve as potential deployment points for the aerial robot. The second metric focuses on graph sharing and the demonstration of continued aerial exploration when the ground robot is unable to follow a low-confidence path.

\subsection{Deployment Zone Identification}\label{subsec:deployment_identification}

\begin{figure}[h]
    \centering
        \includegraphics[width=\linewidth]{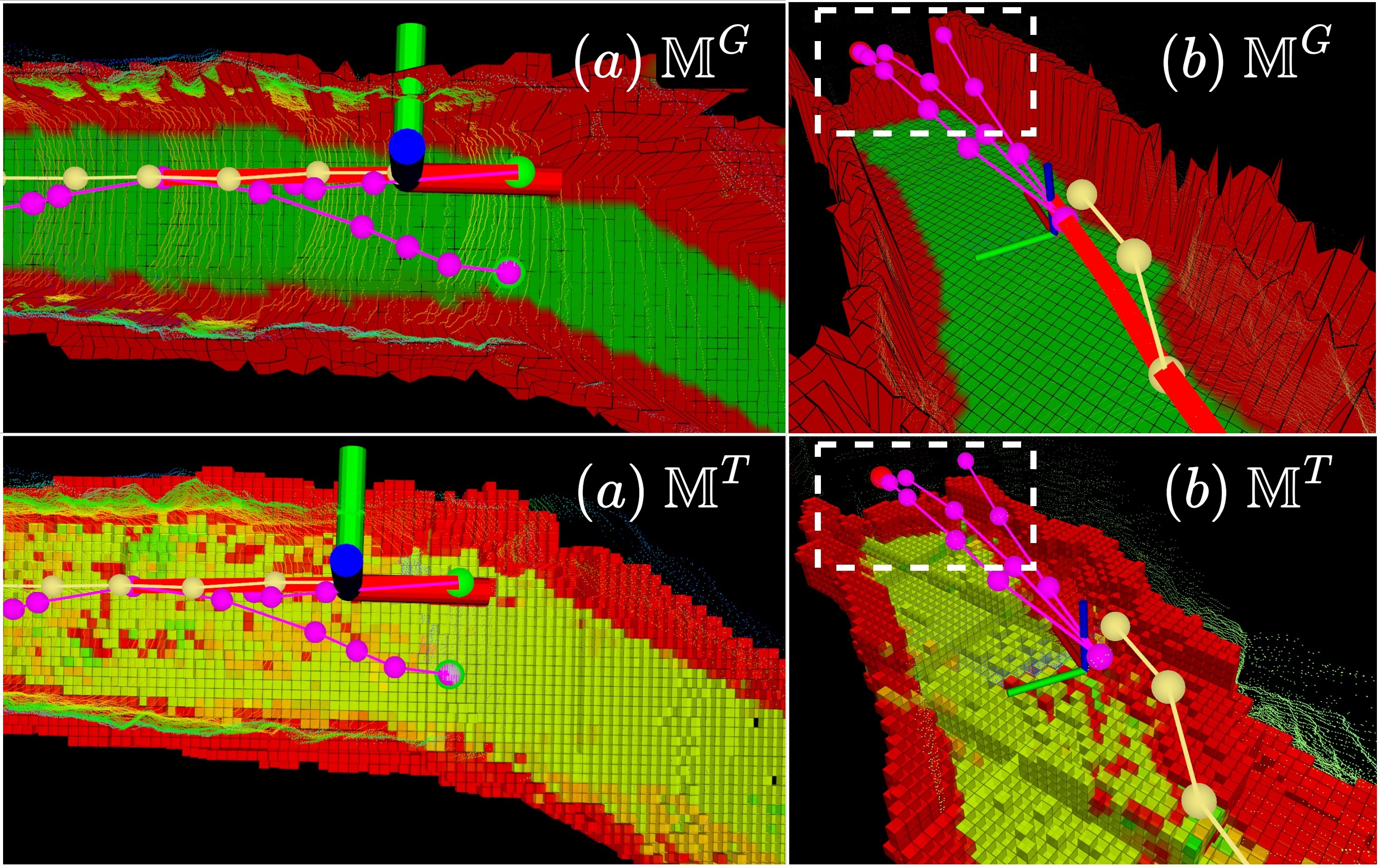}
    \caption{Instances of geometric and semantic traversability maps ($\mathbb{M}^{G} \& \mathbb{M}^{T}$) during exploration}
    \label{fig:trav_instances}
\end{figure}

The experimental evaluation was conducted in a real subterranean tunnel environment featuring a multi-branched topology, cluttered obstacles, and stairways, showcasing verticality. Fig. \ref{fig:trav_instances} illustrates instances of the multi-modal traversability maps ($\mathbb{M}^{G}$ and $\mathbb{M}^{T}$) constructed by the ground robot during autonomous exploration. In Fig. \ref{fig:trav_instances}, $\mathbb{M}^{G}$ represents a 2.5D traversability grid map where green cells indicate high geometric traversability (close to 1) and red cells indicate low traversability (close to 0). Meanwhile, $\mathbb{M}^{T}$ is a 3D voxel map where all visualized voxels are occupied, but the red voxels indicate low traversability and the yellow-green voxels indicate high semantic traversability.

Figure \ref{fig:trav_instances}(a) shows a scenario where the ground robot follows a high-confidence path (green frontier). In contrast, Fig. \ref{fig:trav_instances}(b) depicts a situation where the ground robot encounters a frontier node (red frontier) with very low traversability, resulting in a low-confidence path—a potential deployment point for aerial exploration.

\begin{figure}[h]
    \centering
        \includegraphics[width=\linewidth]{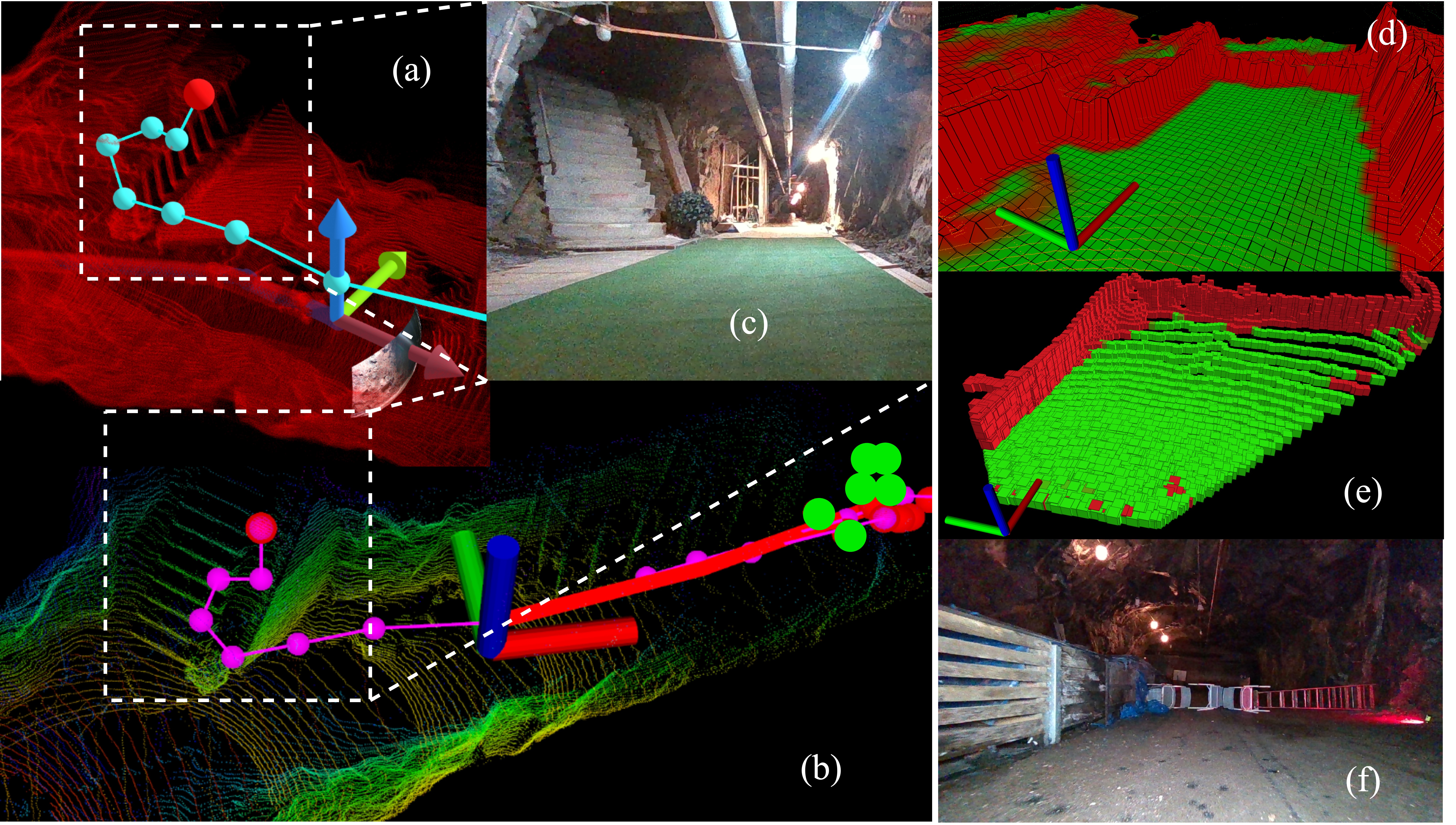}
    \caption{Instances of potential aerial deployment zone identification, demonstrating the role of geometric and semantic traversability in computing the robot's confidence to navigate challenging terrain.}
    \label{fig:stairs_and_junction}
\end{figure}

As the ground robot explores the environment, it encounters vertical obstacles such as steep stairs. In classical exploration approaches, these scenarios would often result in the stairs being misinterpreted as walls, as all occupied voxels in the occupancy map are treated as obstacles. However, the proposed approach addresses this issue by considering both volumetric information gain and traversability analysis when computing the confidence score, allowing the robot to handle such situations more effectively.


\subsection{Integrated Exploration Mission}\label{subsec:integrated_mission}

In addition to identifying potential aerial deployment zones through confidence evaluation, we demonstrate fully autonomous ground-aerial collaborative exploration using the proposed hierarchical graph-based approach. At the deployment zone, the ground robot shares its high-level understanding of the environment by encoding this information into a graph, enabling the aerial robot to continue exploration beyond the untraversable barrier. At the start of the autonomous mission, the ground robot shares its current LiDAR scan with the aerial robot. Utilizing our previous work on a 3D point-cloud map merging framework \cite{stathoulopoulos2024frame}, the aerial robot compares its LiDAR scan with that of the ground robot to compute the static transform between the two. This transform allows the aerial robot to localize itself within the ground robot's global coordinate frame. 

Figures \ref{fig:experiment_collage}(a) to (f) illustrate the various phases of the autonomous exploration mission. The blue point cloud represents the map explored by the ground robot, while the magenta point cloud corresponds to the extended exploration performed by the aerial robot. In Fig. \ref{fig:experiment_collage}(a), the ground robot's local exploration planning is depicted, with green nodes representing traversable frontiers, purple paths denoting candidate paths, and the red path highlighting the final exploration route. Figures \ref{fig:experiment_collage}(b) and (g) capture the moment when the ground robot identifies a potential deployment zone due to the presence of an obstacle preventing further local exploration.

At this point, the ground robot updates the unified graph by incorporating the candidate path leading to the untraversable area, along with the global frontiers. Leveraging the ROS2 action client-server architecture, the ground robot, acting as an action client, prompts the aerial robot to accept the unified graph. Upon receiving the unified graph, the aerial robot sends feedback to the ground robot, acknowledging the request to deploy from the carrier platform. Once the aerial robot has identified the unified frontier, it continues exploration using the same graph-based strategy, as demonstrated in Fig. \ref{fig:experiment_collage}(c). 

The phases shown in Fig. \ref{fig:experiment_collage}(d) through (f) illustrate the continuation of aerial exploration. Since the proposed approach only requires the ground robot to share a graph-based representation of its understanding of the explored environment, the aerial robot can proceed with exploration using the unified graph without the need to share full map data. A clear demonstration of this advantage is provided in Fig. \ref{fig:experiment_collage}, where the aerial robot's graph is derived directly from the ground robot's graph. As a result, the aerial robot's graph contains only valid frontiers corresponding to areas that the ground-aerial robotic system has not yet explored. For more detailed visualization of the full integrated exploration mission, we kindly ask the reviewer to watch the video demonstration of the mission at \url{https://sites.google.com/view/complementary-marsupial-explor/home}.

\section{Conclusion}

In this paper we introduces a hierarchical graph structure that encodes multi-modal traversability and frontier information, allowing for more comprehensive path evaluation during autonomous exploration. Traditional frameworks often struggle with complex terrain features like vertical surfaces and clutter, and collaboration between ground and aerial robots is hindered by inefficient information sharing, often relying on full map exchanges. In response, we propose a robust confidence metric accounts for traversability analysis and predicted volumetric gain facilitating collaborative exploration. By introducing the candidate graph as a sparser layer, the framework facilitates efficient information sharing between robots, reducing the need for full map exchange while guiding exploration toward unified frontier nodes. This not only improves decision-making but also enables dynamic aerial deployment in regions beyond the ground robot’s reach, enhancing collaboration and exploration efficiency in challenging subterranean environments.

\subsection*{--- Open-Source Code Availability}\label{subsec:code}

A key aspect of developing this fully autonomous ground-aerial exploration framework was the integration of open-source modules into ROS2 for multi-agent deployment, as well as the incorporation of geometric and semantic traversability into occupancy maps. Building on Voxblox as the baseline for volumetric mapping, we have extended its functionality by integrating traversability directly into TSDF \& ESDF layers. The ROS2 code for this implementation will be made publicly available at \url{https://github.com/LTU-RAI/TravBlox.git}. 


\addtolength{\textheight}{-11.7cm}

\bibliographystyle{ieeetr}

\bibliography{References}

\begin{thebibliography}{10}

\bibitem{chung2023into}
T.~H. Chung, V.~Orekhov, and A.~Maio, ``Into the robotic depths: analysis and insights from the darpa subterranean challenge,'' {\em Annual Review of Control, Robotics, and Autonomous Systems}, vol.~6, pp.~477--502, 2023.

\bibitem{ebadi2023present}
K.~Ebadi, L.~Bernreiter, H.~Biggie, G.~Catt, Y.~Chang, A.~Chatterjee, C.~E. Denniston, S.-P. Desch{\^e}nes, K.~Harlow, S.~Khattak, {\em et~al.}, ``Present and future of slam in extreme environments: The darpa subt challenge,'' {\em IEEE Transactions on Robotics}, 2023.

\bibitem{dang2020graph}
T.~Dang, M.~Tranzatto, S.~Khattak, F.~Mascarich, K.~Alexis, and M.~Hutter, ``Graph-based subterranean exploration path planning using aerial and legged robots,'' {\em Journal of Field Robotics}, vol.~37, no.~8, pp.~1363--1388, 2020.

\bibitem{dharmadhikari2020motion}
M.~Dharmadhikari, T.~Dang, L.~Solanka, J.~Loje, H.~Nguyen, N.~Khedekar, and K.~Alexis, ``Motion primitives-based path planning for fast and agile exploration using aerial robots,'' in {\em 2020 IEEE International Conference on Robotics and Automation (ICRA)}, pp.~179--185, IEEE, 2020.

\bibitem{yang2021graph}
F.~Yang, D.-H. Lee, J.~Keller, and S.~Scherer, ``Graph-based topological exploration planning in large-scale 3d environments,'' in {\em 2021 IEEE International Conference on Robotics and Automation (ICRA)}, pp.~12730--12736, IEEE, 2021.

\bibitem{patel2024stage}
A.~Patel, M.~A.~V. Saucedo, C.~Kanellakis, and G.~Nikolakopoulos, ``Stage: Scalable and traversability-aware graph based exploration planner for dynamically varying environments,'' in {\em 2024 IEEE International Conference on Robotics and Automation (ICRA)}, pp.~5949--5955, 2024.

\bibitem{lindqvist2022multimodality}
B.~Lindqvist, S.~Karlsson, A.~Koval, I.~Tevetzidis, J.~Halu{\v{s}}ka, C.~Kanellakis, A.-a. Agha-mohammadi, and G.~Nikolakopoulos, ``Multimodality robotic systems: Integrated combined legged-aerial mobility for subterranean search-and-rescue,'' {\em Robotics and Autonomous Systems}, vol.~154, p.~104134, 2022.

\bibitem{de2022marsupial}
P.~De~Petris, S.~Khattak, M.~Dharmadhikari, G.~Waibel, H.~Nguyen, M.~Montenegro, N.~Khedekar, K.~Alexis, and M.~Hutter, ``Marsupial walking-and-flying robotic deployment for collaborative exploration of unknown environments,'' in {\em 2022 IEEE International Symposium on Safety, Security, and Rescue Robotics (SSRR)}, pp.~188--194, IEEE, 2022.

\bibitem{tranzatto2022cerberus}
M.~Tranzatto, T.~Miki, M.~Dharmadhikari, L.~Bernreiter, M.~Kulkarni, F.~Mascarich, O.~Andersson, S.~Khattak, M.~Hutter, R.~Siegwart, {\em et~al.}, ``Cerberus in the darpa subterranean challenge,'' {\em Science Robotics}, vol.~7, no.~66, p.~eabp9742, 2022.

\bibitem{butzke20153}
J.~Butzke, A.~Dornbush, and M.~Likhachev, ``3-d exploration with an air-ground robotic system,'' in {\em 2015 IEEE/RSJ International Conference on Intelligent Robots and Systems (IROS)}, pp.~3241--3248, IEEE, 2015.

\bibitem{shen2017collaborative}
C.~Shen, Y.~Zhang, Z.~Li, F.~Gao, and S.~Shen, ``Collaborative air-ground target searching in complex environments,'' in {\em 2017 IEEE International Symposium on Safety, Security and Rescue Robotics (SSRR)}, pp.~230--237, IEEE, 2017.

\bibitem{fankhauser2014robot}
P.~Fankhauser, M.~Bloesch, C.~Gehring, M.~Hutter, and R.~Siegwart, ``Robot-centric elevation mapping with uncertainty estimates,'' in {\em Mobile Service Robotics}, pp.~433--440, World Scientific, 2014.

\bibitem{chilian2009stereo}
A.~Chilian and H.~Hirschm{\"u}ller, ``Stereo camera based navigation of mobile robots on rough terrain,'' in {\em 2009 IEEE/RSJ International Conference on Intelligent Robots and Systems}, pp.~4571--4576, IEEE, 2009.

\bibitem{saucedoEAT}
M.~A. Saucedo, A.~Patel, C.~Kanellakis, and G.~Nikolakopoulos, ``Eat: Environment agnostic traversability for reactive navigation,'' {\em Expert Systems with Applications}, vol.~244, p.~122919, 2024.

\bibitem{chen2023direct}
K.~Chen, R.~Nemiroff, and B.~T. Lopez, ``Direct lidar-inertial odometry: Lightweight lio with continuous-time motion correction,'' in {\em 2023 IEEE international conference on robotics and automation (ICRA)}, pp.~3983--3989, IEEE, 2023.

\bibitem{oleynikova2017voxblox}
H.~Oleynikova, Z.~Taylor, M.~Fehr, R.~Siegwart, and J.~Nieto, ``Voxblox: Incremental 3d euclidean signed distance fields for on-board mav planning,'' in {\em 2017 IEEE/RSJ International Conference on Intelligent Robots and Systems (IROS)}, pp.~1366--1373, IEEE, 2017.

\bibitem{stathoulopoulos2024frame}
N.~Stathoulopoulos, B.~Lindqvist, A.~Koval, A.-A. Agha-Mohammadi, and G.~Nikolakopoulos, ``{FRAME: A Modular Framework for Autonomous Map Merging: Advancements in the Field},'' {\em IEEE Transactions on Field Robotics}, vol.~1, pp.~1--26, 2024.

\end{thebibliography}

\end{document}